\documentclass{article} 
\usepackage{iclr2021/iclr2021_conference,times}
\usepackage{mathtools}
\usepackage{enumerate}

\usepackage{caption}
\usepackage{amsthm}

\usepackage{subcaption}
\usepackage{graphicx}

\usepackage{natbib}

\usepackage{amsfonts,amsmath,amssymb}
\usepackage{bbm}
\usepackage{xspace}

\usepackage{multicol}
\usepackage{enumitem}

\newcommand{\iid}{\overset{iid}{\sim}}

\providecommand{\mc}[1]{\mathcal{#1}}

\providecommand{\mbb}[1]{\mathbb{#1}}

\newcommand{\argmax}{\operatornamewithlimits{argmax}}

\usepackage{subcaption}

\usepackage{hyperref}
\usepackage{url}

\title{Inducing a hierarchy \\for multi-class classification problems}

\author{Hayden~S.~Helm$^{*}$ \\
Microsoft Research \& Johns Hopkins University \\
\texttt{haydenshelm@gmail.com}
\AND
Weiwei~Yang
\& Sujeeth~Bharadwaj
\& Kate~Lytvynets
\& Oriana~Riva
\& Christopher~White \\
Microsoft Research \\
\texttt{\{weiwya, 
sujeeth.bharadwaj, 
kalytv, Oriana.Riva,
chwh\}@microsoft.com} \\
\AND
Ali~Geisa \& Carey~E.~Priebe \\
Johns Hopkins University \\
\texttt{\{ageisa1, cep\}@jhu.edu}
}

\newcommand\blfootnote[1]{%
  \begingroup
  \renewcommand\thefootnote{}\footnote{#1}%
  \addtocounter{footnote}{-1}%
  \endgroup
}

\iclrfinalcopy 
\begin{document}
\maketitle
\blfootnote{
$ ^{*} $ corresponding author
}

\begin{abstract}
In applications where categorical labels follow a natural hierarchy, classification methods that exploit the label structure often outperform those that do not. Unfortunately, the majority of classification datasets do not come pre-equipped with a hierarchical structure and classical ``flat" classifiers must be employed. In this paper, we investigate a class of methods that \textit{induce} a hierarchy that can similarly improve classification performance over flat classifiers. The class of methods follows the structure of first clustering the conditional distributions and subsequently using a hierarchical classifier with the induced hierarchy. We demonstrate the effectiveness of the class of methods both for discovering a latent hierarchy and for improving accuracy in principled simulation settings and three real data applications.
\end{abstract}






\section{Introduction}
Machine learning practitioners are often challenged with the task of classifying an object as one of tens or hundreds of classes. To address these problems, algorithms originally designed for binary or small multi-class problems are applied and naively deployed. In some instances the large set of labels comes pre-equipped with a hierarchical structure -- that is, some labels are known to be mutually semantically similar to various degrees. In these cases, algorithms have been developed to leverage the structure to improve classification performance \citep{silla2011survey}. Unfortunately, the majority of large multi-class classification problems do not come with structured labels and the potential performance gains realized for hierarchical problems are unavailable. 

In this note we investigate the effect of assuming the existence of a latent hierarchy even when one is not explicitly specified in the set of labels. We demonstrate that this assumption, coupled with inducing a hierarchy via clustering the conditional distributions, is beneficial in a large regime of a Gaussian-based simulation setting and in multiple real data settings. Our results suggest that for large multi-class problems, inducing a hierarchy is an effective pre-processing step when using classifiers out-of-the-box.

\noindent \textbf{Contribution: We present evidence that hierarchical classifiers should be strongly considered in large multi-class classification problems -- both when there exists a latent or known hierarchy \textit{and} when there does not.} 
\label{introduction}

\section{Background}
We first provide background and notation related to statistical pattern recognition before describing hierarchical classification as an extension of the classical setting. We then discuss different ways to cluster conditional distributions, how to measure the quality of clusters when a ground truth is available, and contextualize our work within the broader literature.

\subsection{Statistical Pattern Recognition}
Let 
\begin{align*} (X, Y), (X_{1}, Y_{1}), \ldots, (X_{n}, Y_{n}) \iid  F 
\end{align*}
be random variables distributed according to the joint distribution $ F $ with input realizations $ X_{i} = x_{i} \in \mc{X} $ and label realizations $ Y_{i} = y_{i} \in \{1, \hdots, k\} $. We let $ F_{X} $ be the marginal distribution of $ X_{i} $.  The objective in statistical pattern recognition \citep{devroye2013probabilistic, duda2012pattern} is to use the training data $ \mc{D}_{n} = \{(X_{i}, Y_{i})\}_{i \in \{1, \hdots, n\}} $ to learn a decision function $ h_{n} $ that correctly maps $ X $ to the true but unknown $ Y $. 

Formally, the objective in statistical pattern recognition is to minimize the risk of $ h_{n} $. The risk of a decision function $ h: \mc{X} \to \{1, \hdots, k\} $
is defined as the expected value of a loss function $ \ell: \{1, \hdots, k\} \times \{1, \hdots, k\} \to [0, \infty) $ evaluated at the output of the decision rule and the truth: $ \ell(h(X), Y) $. Or, the risk $ R $ of $ h $ for $ F $ is
\begin{align*}
    R_{F}(h) := \mbb{E}_{F}\left[\ell(h(X), Y)\right].
\end{align*} 

The \textit{Learning Efficiency} (LE) \citep{vogelstein2020general} of $ h_{n} $ relative to $ h'_{n} $ for a given $ n $ is the ratio of the corresponding risks:
\begin{align}\label{eq:learning-efficiency}
    LE(h_{n}, h'_{n}) = \frac{R_{F}(h_{n})}{R_{F}(h'_{n})}.
\end{align} The decision rule $ h'_{n} $ is preferred to $ h_{n} $ for a given $ n $ if $ LE(h_{n}, h'_{n}) > 1 $.  

\subsubsection{Hierarchical classification}
Hierarchical classification \citep{gordon1987review, silla2011survey} is an extension of classical statistical pattern recognition. For notational and conceptual simplicity, we focus on the case where the hierarchy is a tree, though the methods developed in this paper generalize to other hierarchical structures including directed acyclic graphs.

Let 
\begin{align*} 
(X, Y^{1}, \hdots, Y^{j}), \\
\; (X_{1}, Y^{1}_{1}, \; \ldots, Y^{j_{1}}_{1}), \; \ldots, (X_{n}, Y^{1}_{n}, \; \ldots, Y^{j_{n}}_{n}) \iid  F_{hier.}
\end{align*} be random variables distributed according to the joint distribution $ F_{hier.} $. In hierarchical classification each pattern $ X $ has associated with it $ j $ different labels $ Y^{1}, Y^{2}, \hdots, Y^{j} $. The label $ Y^{1} $ is a random variable with realizations in a set of labels unique to the first level of the hierarchy, the label $ Y^{2} $ is a random variable with realizations in a set of labels unique to the second level conditioned on $ Y^{1} $, and so on.

\begin{figure*}
    \centering
    \includegraphics[width=\linewidth]{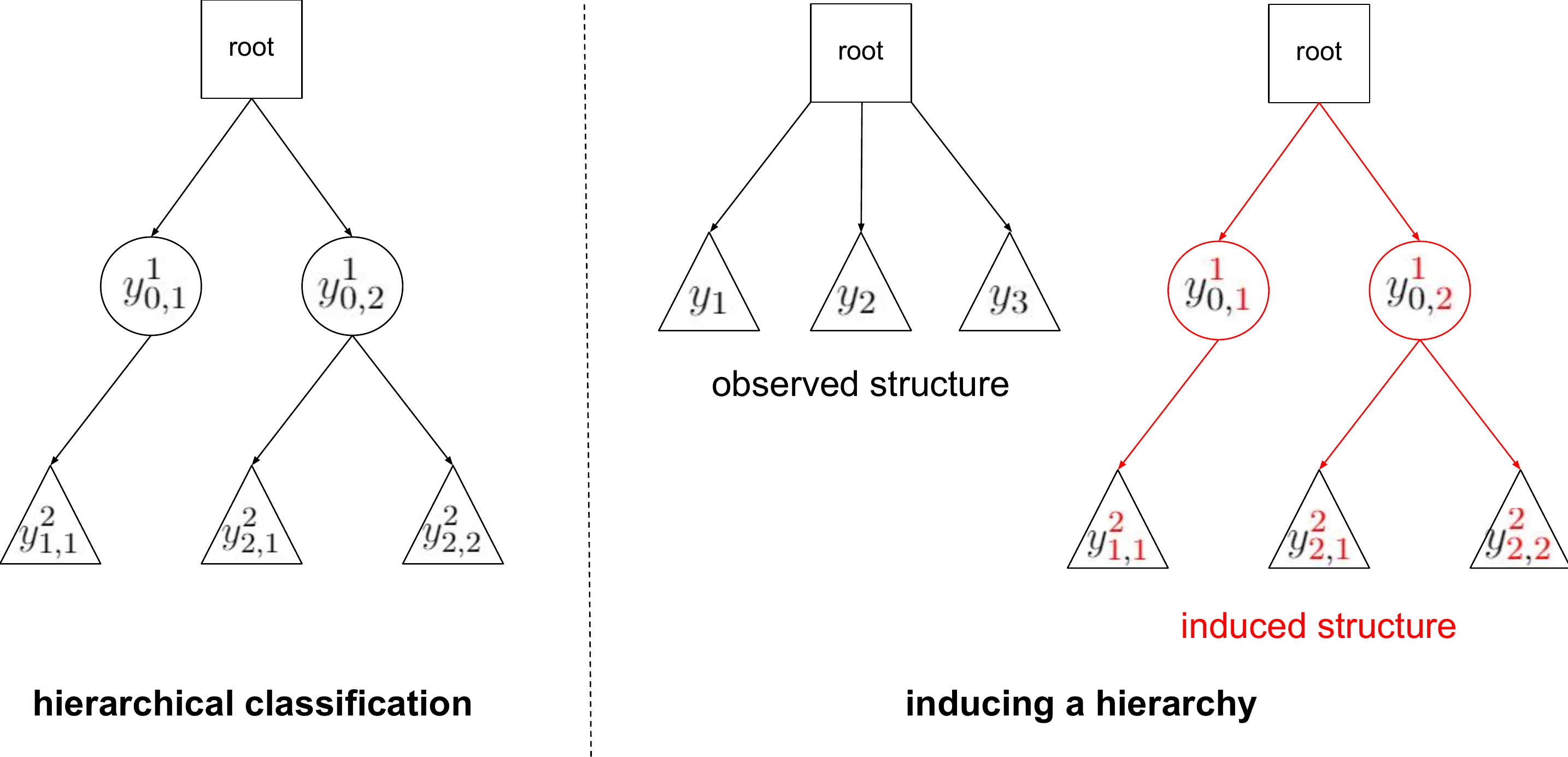}
    \caption{Illustrations of the structure that is typically observed in hierarchical classification (left) and what is observed and can be induced in settings that do not come pre-equipped with a hierarchy (right). Inducing a hierarchy requires clustering conditional distributions. Once a hierarchy is induced, hierarchical classification methods can be used.}
    \label{fig:hierarchical-classification}
\end{figure*}

Notationally, we describe a label using three scripts -- a superscript on the label corresponds to the level of the hierarchy, the first subscript corresponds to the index of the the parent label, and the second subscript corresponds to the index of the label for the current level. Since we focus only on tree hierarchical structures, every pattern where $ y^{j} \in \{y^{j}_{\ell,1}, \hdots, y^{j}_{\ell,k}\} $ is such that $ y^{j-1} = y^{j-1}_{\ell', \ell} $. Figure \ref{fig:hierarchical-classification} shows an example of a set of labels with a hierarchical structure. 

Note that even though each path to a ``leaf'' of the hierarchy may not be the same length, we can still ascribe a label signature $ \vec{Y} = (Y^{1}, Y^{2}, \hdots, Y^{j_{max}}) $ to each pattern by letting $ Y^{\ell} = 0 $ for all $ \ell > j $. Conditioned on a label signature $ \vec{Y} $, all patterns with the same label signature are $ i.i.d. $ according to $ F_{\vec{Y}} $. 

There are many possible inference tasks within the hierarchical classification setting, including multi-label classification \citep{cerri2014hierarchical, chen2019deep} and just-the-leaves classification \citep{bi2014mandatory}. We consider a setting where the hierarchy is not known \textit{a priori} and thus focus on just-the-leaves classification. As in classical statistical pattern recognition, the goal in just-the-leaves classification is to use a decision rule $ h $ that minimizes the risk $ R_{F}(h) $. 

Decision rules that attempt to leverage the additional structure present in hierarchical classification fall into three categories: flat, global, and local. Flat decision rules ignore the hierarchical structure and perform multi-class classification. Global decision rules are a single multi-class decision rule that use the hierarchical structure explicitly when training. Local decision rules construct decision functions at various points within the hierarchy. We refer the interested reader to \citep{silla2011survey} for a more detailed discussion on the different types of hierarchical decision rules and ways to train them. The approach we consider herein is a local decision rule.

We let the maximum length of a label signature be 2, i.e. $ j_{max} = 2 $, for simplicity. That is, each leaf-level label has at most one parent label. We often refer to parent labels as coarse labels and leaf-level labels as fine labels. 

\subsection{Clustering distributions}
In our setting we assume there exists a hierarchy but that we only observe the leaf-level class labels (i.e. the label signature for $ X $ is unknown except for $ Y^{j}$). One way to attempt to recover the latent hierarchy is to cluster the conditional distributions $ \mc{F} = \{F_{X|Y=1}, F_{X|Y=2}, \hdots, F_{X|Y=k}\} $\footnote{In cases where $ j_{max} > 2 $ one can recursively cluster the conditional distributions as in \citet{lyzinski2016community} to estimate the entire label signature.}. We let $ \mc{F}_{X} $ be the set of conditional distributions on the pattern space $ \mc{X} $ and let $ \mc{F} \subseteq \mc{F}_{X} $.

Depending on the assumptions on $ \mc{F} $, there are many different ways to cluster the conditional distributions. For example, if we assume that the elements of $ \mc{F} $ are parametric $ \mc{F} = \{F_{\theta}: \theta \in \Theta\} $ then we can simply cluster the parameter vectors. Letting $ \theta_{F} = \mbb{E}_{F}(X) $ and $ X \in \mbb{R}^{d} $ then any clustering method on $ \mbb{R}^{d} $, such as K-Means \citep{1017616} or Gaussian Mixture Modeling via the Expectation-Maximization algorithm (GMM) \citep{reynolds2009gaussian}, can be used to induce a hierarchy. Theoretical guarantees for the clusters depends on further structure on the Type II distribution \citep{good1965estimation}, or the distribution of the parameter vectors \citep{lyzinski2014perfect}. 

More general clustering techniques for distributions involve mapping the patterns to a high-dimensional space via a feature map $ \phi: \mc{X} \to \mc{H}$ and calculating the expectation $ \mbb{E}_{F}\left(\phi(X)\right)$ in $ \mc{H} $. With an appropriately chosen $ \phi $, the expectation $ \mbb{E}_{F}\left(\phi(X)\right) $ is unique to $ F $ \citep{berlinet2011reproducing, gretton2012kernel}. Hence, the hierarchical structure induced by clustering the expectations in $ \mc{X} $ is not necessarily the same hierarchical structure induced by clustering the expectations in $ \mc{H} $.

In some cases the appropriate feature map is unknown or $ \mc{H} $ is infinite and $ \mbb{E}_{F}(\phi(X)) $ cannot be calculated directly. In these cases, (dis)similarity-based clustering techniques can be used. For example, associated with some feature maps $ \phi $ are kernels $ \kappa: \mc{X} \times \mc{X} $ that measure the (dis)similarity between two patterns in $ \mc{X} $. In cases where $ \phi $ induces a kernel, differences involving average measurements in $ \mc{X} \times \mc{X} $ (i.e., $\mbb{E}_{X, X' \sim F}\left(\kappa(X, X')\right) $) are equivalent to (dis)similarities on distributions in $ \mc{H} $. Hence, without knowing the feature map it is still possible to calculate the (dis)similarities between distributions in $ \mc{H} $. This is exactly the idea behind the ``kernel trick" \citep{scholkopf2001kernel}, Maximum Mean Discrepancy (MMD) \citep{gretton2012kernel}, and recent ideas in statistical hypothesis testing \citep{doi:10.1080/01621459.2018.1543125}.

Once pairwise (dis)similarities between conditional distributions are measured there are two options for inducing a hierarchical structure. The first is to cluster the conditional distributions directly using the pairwise similarity matrix via methods like DBSCAN \citep{Ester96adensity-based}. The second is to first embed the pairwise similarity matrix so that the conditional distributions are represented as points in $ \mbb{R}^{d} $ and then to deploy the methods above. Embedding methods such as multi-dimensional scaling \citep{wickelmaier2003introduction} and spectral embedding \citep{von2007tutorial} are well studied in particular settings \citep{dokmanic2015euclidean, athreya2017statistical} and are generally used as dimensionality reduction techniques for clustering \citep{lyzinski2014perfect} and classification \citep{tang2013universally}, among other tasks.

\subsubsection{Task similarity as a similarity on conditional distributions}\label{subsubsec:tasksim}

Recent advancements in transfer learning \citep{pan2009survey} have inspired definitions of similarities on classification distributions \citep{gao2020informationgeometric, helm2020partitionbased}. In our setting we deal with conditional distributions, not classification distributions, though we can still deploy the novel similarities by introducing a reference conditional distribution and artificially creating two classification distributions. Herein we study task similarity \citep{helm2020partitionbased} as a similarity used to induce a hierarchy. 

In particular, to measure the similarity $ s(F, G) $ between two conditional distributions $ F $ and $ G $ using task similarity $ TS $, we introduce a third distribution $ H $ and define two classification distributions: $ F' = p(X | Y \sim F) F + (1 - p(X | Y \sim F)) H $ and $ G' = p(X | Y \sim G)G + (1 - p(X | Y \sim G)) H $. We then let $ s(F, G) = \frac{1}{2}\left(TS(F', G') + TS(G', F') \right)$. 

In large multi-class problems we randomly sample $ H $ from the set conditional distributions with $ F $ and $ G $ removed. In practice, we randomly sample multiple $ H_{i} $ and use the average of the $ s(F, G) $ as the similarity that we use to construct a pairwise similarity matrix. We then embed the similarity matrix and cluster the conditional distributions as above. 

We note that using a reference distribution to define a (dis)similarity is not unique to task similarity. The information theoretic Jenson-Shannon divergence \citep{dagan1997similarity, manning1999foundations} uses $ H = \frac{1}{2}\left(F + G\right)$ as a reference distribution when measuring relative entropy.

In this paper we investigate two different clustering methods for inducing a hierarchy: estimating the conditional means and clustering them via GMM and measuring the pairwise task similarity between conditional distributions and using a composition of spectral embedding and GMM.



\subsubsection{Measuring the quality of clusters}

Given a partition of the conditional distributions, there are many ways to measure the quality of the partition, including the Rand index \citep{doi:10.1080/01621459.1971.10482356} and variants, normalized mutual information and variants \citep{romano2014standardized}, variation of information \citep{10.1007/978-3-540-45167-9_14}, etc. Generally, there is no measure of similarity between partitions that is both finite and satisfies a set of desirable properties outlined in \citep{MEILA2007873}. Hence, the choice of measure is a bit arbitrary. 

When we can compare the induced hierarchy to a true-but-unknown hierarchy in our experiments we use the adjusted Rand index (ARI) \citep{doi:10.1080/01621459.1971.10482356} because it is both bounded and adjusted to take into account the performance of a random clustering.  

\subsection{Related Work}
Inferring or inducing a hierarchy has been studied in the contexts of document classification \citep{punera2005automatically}, general multi-label classification \citep{brucker2011multi}, and leaf-only classification settings where distributional assumptions are made \citep{punera2006automatic}. 

Most similar to our work is work by Punera and colleagues \citep{punera2005automatically, punera2006automatic}. They first propose a method to induce a hierarchy with a binary tree structure via intelligently initialized K-Means \citep{punera2005automatically} clustering of the class conditional means. In later work they propose inducing a hierarchy via first measuring the Jensen-Shannon distances between the conditional distributions and then clustering the distance matrix using agglomerative clustering. The agglomerative clustering approach allows for more general tree structures that can potentially find semantically meaningful structure higher in the hierarchy.

In our setting we observe two sets of patterns from continuous distributions, $ \{x_{1}, \hdots, x_{n}\} $ and $ \{x'_{1}, \hdots, x'_{m}\} $, such that (typically) for all $ i \in \{1, \hdots, n\} $ it is such that $ x_{i} \neq x'_{j} $ for all $ j \in \{1, \hdots, m\} $. Hence, divergences and distances inspired by discrete information theory, such as the Kullback-Leibler divergence and Jenson-Shannon distance, are uninformative for each pair of empirical distributions. Mitigating this requires making distributional (e.g., Gaussianity) assumptions for each of the conditional distributions such that the support for the distributions have a non-empty intersection. Distributional assumptions are problematic in high dimensions where estimating covariance structures with little data is infeasible or unstable.

\label{background}

\section{Inducing a hierarchy}

To induce a hierarchical structure on the labels and to ultimately make leaf-level predictions for unlabeled observations in $ \mc{X} $ we compose a clustering technique and a hierarchical classifier. 



For both the simulation and real data settings we use a sequence of uncertainty forests \citep{guo2019estimating}, a forest-based method designed to consistently estimate the class-conditional probability vector $ \left[p(y_{1}|x), \hdots, p(y_{k}|x)\right]^{T} $ as in \citep{chen2019deep}. In particular, the probability that a pattern $ x $ has leaf-level label $ y $ is given by the chain rule of conditional probabilities implied by its label signature. Assuming a two-level tree, the probability that a pattern is from class $ y^{2}_{\ell, \ell'} $ can be written as 
\begin{align}\label{eq:bayes}
    p(y^{2}_{\ell, \ell'}|x) = p(y^{2}_{\ell, \ell'}| y^{1}_{0, \ell'}, x) \cdot p(y^{1}_{0, \ell'} | x).
\end{align}
Hence, the leaf-level conditional probability $ p(y^{2}_{\ell, \ell'} | x) $ can be estimated using a collection of uncertainty forests\footnote{Generalizing to directed acyclic graphs requires including a summation over the first-level labels. Generalizing to hierarchies with path lengths greater than two requires applying \eqref{eq:bayes} recursively.} -- one for the first level of the hierarchy, i.e., $ p(y^{1}_{0, \ell'}) $, and then one for each of the classes in the first level, i.e., $ p(y^{2}_{\ell, \ell'}) $. The hierarchical classifier then simply predicts the $ \argmax $
amongst leaf-level labels. 

In each setting we compare the hierarchical classifiers to a flat classifier, via learning efficiency \eqref{eq:learning-efficiency}, with the risk of the flat classifier always in the denominator. A learning efficiency greater than 1 implies that for the given data the hierarchical approach is preferred to the flat approach. 

\label{method}

\subsection{Simulations}

\begin{figure*}[t]
    \centering
    \begin{subfigure}{\textwidth}
        \centering
        \includegraphics[width=\linewidth]{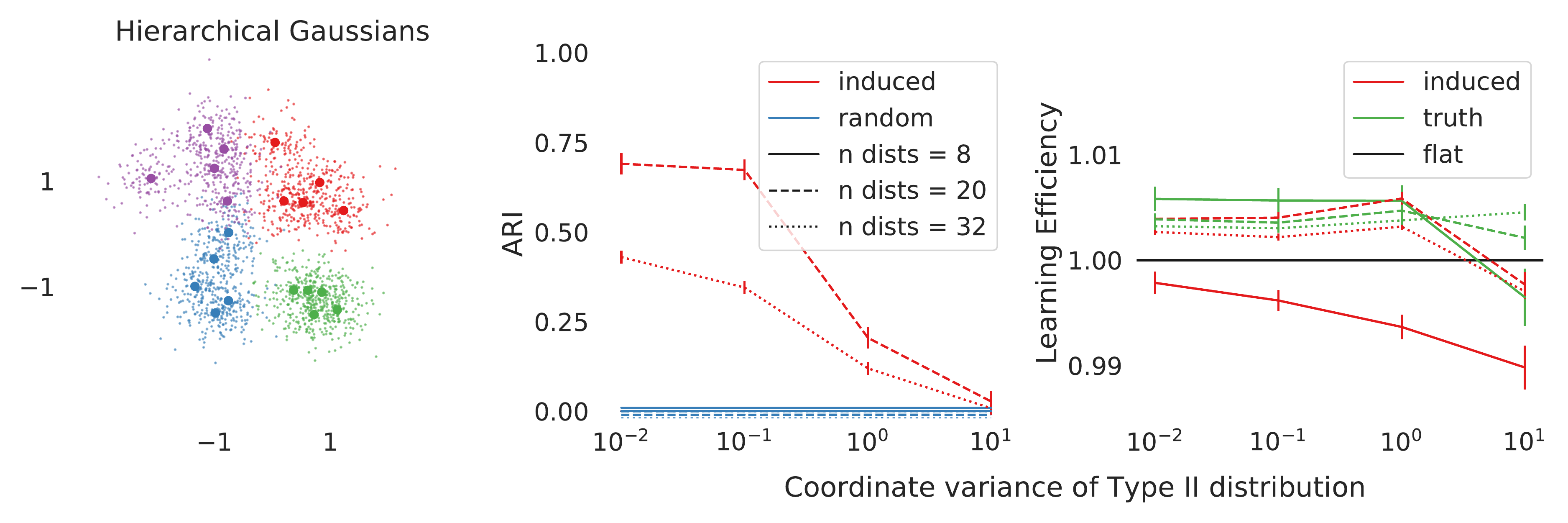}
    \end{subfigure}
    \caption{Hierarchical classification can significantly outperform flat classification in regimes where the variance corresponding to the parent label is smaller than the variane corresponding to the fine label. In particular, when a latent hierarchy exists this structure can be inferred and used for hierarchical classification to improve performance in particular parameter settings. The error bars on the right two figures represent standard errors of the corresponding averages.}
    \label{fig:simulations}
\end{figure*}

We study the effect of inducing hierarchical structure in a Gaussian-based simulation setting. In our setting the conditional distributions are parameterized by mean vectors that are sampled from a mixture of $ 4 $ Gaussians. The mean vectors that are from the same element of the mixture share a parent label in the latent hierarchy. For each experiment we have access to $ 50 $ points sampled from a Gaussian corresponding to each mean vector. 

To cluster the distributions, then, we first estimate the sample means from the observed patterns from each class and then estimate a mixture of Gaussians via GMM. We refer to the mixture of Gaussians for the parent labels as the Type II distribution.


The $ 4 $ Type II Gaussians are centered at each of $ \{(1,1), (1,-1), (-1,1), (-1,-1)\} $ with covariances equal to $ \gamma_{0} \cdot I_{2} $. We sample \{2, 5, 8\} different mean vectors from each of the $ 4 $ Gaussians, resulting in an \{8, 20, 32\}-class classification problem. From each of the sampled mean vectors we then sample 50 patterns from a Gaussian centered at the mean vector with covariance $ \gamma_{1} \cdot I_{2} $. An example of this setting is shown in the left panel of Figure \ref{fig:simulations} with $ \gamma_{0} = 2 $ and $ \gamma_{1} = 1 $. The small dots correspond to the observed patterns, the large dots correspond to the sample means of each the conditional distributions, and the color corresponds to the element of the Type II distribution from which the mean vector is sampled. 

The center panel of Figure \ref{fig:simulations} shows the effect of varying the coordinate variance (parameterized by $ \gamma_{0} $) of the Type II distribution as measured by the ARI between the predicted clusters and the true clusters with $ \gamma_{1} = 1 $ fixed. We compare the ARI of the estimated clusters to the ARI information of a random clustering with the same cluster structure. As $ \gamma_{0} $ increases, the mean vectors from the same element of the mixture are more variable. Thus, the latent structure is harder to recover and the ARI of the estimated hierarchy approaches that of the random clustering.

The right panel of Figure \ref{fig:simulations} shows the effect of varying $ \gamma_{0} $ on the learning efficiency of hierarchical classifiers trained on the induced hierarchy and the true hierarchy. Hierarchical classifiers were trained using $ 10 $ trees for the parent labels and $ 5 $ trees for each class within the parent labels. The flat classifier was trained using $ 10 + 5(4) $ trees, where $ 4 $ is the number of parent labels in the true hierarchy. Each tree in every forest had a maximum depth of $ 20 $. 

In the regime where the Type II variance is smaller than the variance in the conditional distributions, both hierarchical classifiers are more efficient than the flat classifier. Notably, some of the regime where the hierarchical classifiers are more efficient correspond to the regime where the ARI is not particularly high, perhaps indicating that using any hierarchy -- not just the true hierarchy -- can result in a performance gain.

While we do not pursue any theoretical results showing that hierarchical methods can be more efficient that flat methods, we do plan to pursue them in future work in an even simpler setting that is more amenable to analysis. As far as we are aware there are no such results that exist within the literature.

\label{simulations}

\subsection{Cifar, CheXpert, and Icons}

We next study the effect of inducing a hierarchy for classification on three real datasets -- CIFAR~100 \citep{Krizhevsky09learningmultiple}, CheXpert \citep{irvin2019chexpert}, and Icons. For each of the three datasets we train classifiers using a random 10\% of the the data and measure the accuracy of a flat classifier and two pairs of hierarchical classifiers. When applicable we include the accuracy of a hierarchical classifier trained using the true-but-unknown hierarchy.

The hierarchical classifiers differ only in how the label hierarchy is induced. Each pair -- one based on clustering the pairwise task similarity matrix and the other clustering the estimated conditional means -- consists of an induced hierarchy and a random hierarchy that shares the same cluster structure. 

For each hierarchical and flat classifier, we use a collection of uncertainty forests as above, always controlling the number of trees in each forest and the maximum depth of each tree. For the two datasets (CIFAR \& CheXpert) where there exists a true-but-unknown hierarchy we also measure the quality of the induced hierarchy via ARI.

The ARI and accuracy results for the three settings are shown in Table \ref{tab:results}. Detailed descriptions of the experiments are below.

\begin{table*}[!t]
\centering
\begin{tabular}{||c||c c c c c c||}
 \hline
 \textbf{ARI} & Flat & Cond. Mean & Random (CM) & Task Sim. & Random (TS) & Truth \\ [0.5ex] 
 \hline\hline
 CIFAR & 0.00 & 0.40 (0.01) & 0.00 (0.00) & 0.27 (0.01) & 0.00 (0.00) & 1.00  \\
 CheXpert & 0.00 & -0.02 (0.02) & 0.00 (0.00) & 0.09 (0.04) & 0.00 (0.00) & 1.00 \\
 Icons & - & - & - & - & - & -  \\
 \hline
 \end{tabular}
\begin{tabular}{||c||c c c c c c||}
 \hline
 \textbf{Acc.} & Flat & Cond. Mean & Random (CM) & Task Sim. & Random (TS) & Truth \\ [0.5ex] 
 \hline\hline
 CIFAR & 27.1 (0.7) & 51.0 (0.4) & 40.3 (0.7) & \textbf{53.8 (0.3)} & 39.7 (0.7) & 52.9 (0.2)  \\
 CheXpert & 41.8 (0.1) & 41.9 (0.1) & 41.9 (0.1) & 41.9 (0.1) & 41.9 (0.1) & \textbf{42.1 (0.0)} \\
 Icons & 38.6 (0.8) & 73.7 (1.2) & 66.8 (1.5) & \textbf{76.2 (0.1)} & 67.5 (1.0) & -  \\
 \hline
 \textbf{Avg. LE} & 1.00 & 1.61 & 1.36 & 1.72 & 1.37 & - \\
 \hline
 \end{tabular}
 \caption{Average ARIs (top) and accuracies (bottom) of the hierarchical methods in the studied settings. Learning efficiencies correspond to the ratio of the average error of the flat classifier to the average error of the hierarchical classifier averaged over the three datasets. Standard errors, when reported, are in parentheses. Accuracies that are uniquely in a maximum range defined by the standard errors are bolded. Detailed descriptions of the experiments, including data pre-processing, are in the main text.}
 \label{tab:results}
\end{table*}

\subsubsection{CIFAR 100}
We first investigate our problem in the context of CIFAR~100 \citep{Krizhevsky09learningmultiple}.
Recall that CIFAR~100 comes pre-eqipped with a label hierarchy. That is, each image has both a coarse and a fine label. For example, all patterns with the fine label ``lion" are also labeled ``large carnivore". All of the coarse labels are semantically (biologically or otherwise) meaningful -- flowers share a common coarse label, large carnivores share a common coarse label, small mammals share a common coarse label, etc. There are a total of twenty coarse labels with five fine labels each. Each random 10\% of the data that we sample is not stratified. 

For our experiment, we first generate $ d=2,048 $ dimensional representations of the original images using Google's pre-trained BiT-M R101x1 model (\cite{Kolesnikov2020BigT}) without any fine-tuning.  

We then induce a hierarchy in two ways. The first is by clustering the conditional means via GMM as implemented by \texttt{graspologic} \citep{JMLR:v20:19-490}. Recall that GMM estimates a mixture of Gaussians, i.e. a mean vector and a covariance matrix, per cluster of conditional means. In $ d = 2,048 $-space the estimated covariances are ill-defined. To mitigate this, we first project the $ d = 2,048 $ dimensional representations of the images into a $ d=128$-space learned via Principal Component Analysis (PCA) \citep{bro2014principal}. Once the representations of the images are projected we estimate the conditional means and cluster them with the assumption that every cluster has the same covariance structure (to mitigate computational issues).

The second is by measuring the pairwise task similarities, as described in Section \ref{subsubsec:tasksim}, using $ 10 $ randomly sampled reference distributions in our measurement. We then process the pairwise task similarity matrix so that the entries are between 0 and 1. Finally, we embed the processed matrix into a $ d=16$ dimensional space via the adjacency spectral embedding and cluster the embedded distributions using GMM. When repeated 10 times each with a different random sample of 10\% of the data, the clusters from the conditional means outperform the clusters from task similarity as measured by ARI.

Once we have an induced hierarchical structure we then train two uncertainty forests for each hierarchy -- one forest for the coarse labels and one for each label in the set of coarse labels. The final classifier returns the label with the highest posterior probability as described by Eq. \ref{eq:bayes}. In this setting the forest for the coarse label contains $ 100 $ trees with a maximum depth of $ 10 $,  and each of the forests for the fine labels contains $ 100 $ trees with a maximum depth of $ 10 $. As mentioned above, we compare the accuracies of the hierarchical classifiers to a flat classifier and a hierarchical classifier that uses the true latent hierarchy. The flat classifier is a single uncertainty forest with $ 300 $ trees and a maximum depth of $ 20 $. In general, increasing the number of trees for the flat classifier will not significantly increase the accuracy \citep{probst2017tune}. Accuracies were measured using the traditional test set. 

In this setting, the hierarchical approaches significantly improve upon the flat classifier, with learning efficiencies of the average accuracies ranging from 1.21 (from random hierarchy with the same structure as those induced by task similarity) to 1.58 (from hiearchies induced by task similarity). Interestingly, the classifier trained on the hierarchy induced by task similarity outperforms the classifier trained on the true hierarchy. We posit that this result is a function of the true hierarchy being semantically meaningful and not necessarily visually meaningful. For example, the classes ``whale", ``shark" and ``dolphin" do not all share the same true coarse label but often share a coarse label in the induced hierarchies. The same is true for ``rocket", ``road" and ``skyscraper". Our result suggests that the vision-based hierarchies can improve performance upon the true but semantically-based hierarchy for vision tasks.

\subsubsection{CheXpert}
We next investigate the effect of inducing a hierarchy on a chest X-ray dataset CheXpert \citep{irvin2019chexpert}. CheXpert is comprised of $ 224,316 $ frontal and lateral images of $ 65,420 $ patients. For simplicity we only considered frontal images. Each frontal image has a corresponding multi-label indicating whether or not a set of diseases, such as pneumonia and a cardiomegaly, is present in the patient. In the paper and challenge associated with CheXpert the authors outline five different classes that are challenge classes, picked because of their ``clinical improtance and prevalance" -- ``atelectasis", ``cardiomegaly", ``consolidation", ``edema", and ``pleural effusion". 

Since our methods do not naturally handle multi-labels, we only included patients with either only a single disease or patients with only a single challenge disease. This pre-processing left us with approximately $ 90,000 $ images each with a corresponding label in \{``no finding", ``cardiomegaly", ``lung lesion", ``edema", ``consolidation", ``pneumonia", ``atelectasis", ``pneumothorax", ``pleural effusion", ``pleural other", ``fracture"\}. 

As with our CIFAR experiment, we extracted $d = 2048 $ representations of the original images using the pre-trained BiT-M R101x1 without fine-tuning the model. 

To induce a hierarchy via conditional means we sampled a non-stratified 10\% of the data, projected the representations of the images into $ d = 128$-space, estimated the conditional means, and clustered the conditional means via GMM with the assumption that each cluster has the same covariance structure. We induced a hierarchy via the pairwise task similarity matrix using the same pipeline as in CIFAR except using only $ 5 $ reference distributions to measure the task similarities. In this setting the hierarchy induced by task similarity outperforms the hierarchy induced by clustering the conditional means, as measured by ARI. In general, the ARIs in this setting are much smaller than the ARIs in the CIFAR setting. We think that this is mainly due to only considering $ 11 $ classes.

Similarly, the learning efficiencies for the hierarchical approaches are unimpressive compared to the CIFAR results -- all of them, including the one corresponding to the true hierarchy, are effectively $ 1 $. In this setting, the two forests for each hierarchical approach each contained $ 100 $ trees with a maximum depth of $ 10 $. The flat classifier contained $ 300 $ trees with a maximum depth of $ 20 $. For each iteration of the experiment, accuracies were measured using the 90\% of data not selected for training. 

\subsubsection{Icons}

The last setting we consider is based on the Kaggle Common Mobile Web \& App Icons dataset (https://www.kaggle.com/testdotai/common-mobile-web-app-icons). The Icons dataset contains $ 106 $ classes, including $ 105 $ corresponding to a type of icon, such as ``volume", and one corresponding to a ``negative" class. There is no hand labeled hierarchy available for this dataset, though there exists sets of icons that are either semantically or visually related. For example, there are images in the dataset for ``arrow left", ``arrow right", and ``arrow up" icons. 

Before inducing hierarchies and training classifiers, we first fine-tuned the pre-trained BiT-M R101x1 to improve the representations of the images a non-stratified random 10\% of the data. After fine-tuning we removed the data that was used from the remainder of our experiment and further removed data corresponding to classes with fewer than $ 250 $ images -- resulting in a $ 95 $ class classification problem.

We induced hierarchies using both the conditional means-based method and the task similarity-based method described above. We measured the task similarities using $ 5 $ reference distributions. Since there is no hand-labeled hierarchy we cannot report the average ARI of the clusters from each method, though both methods appear to be able to capture visually meaningful relationships. For example, the task similarity-based method discovers the cluster \{``arrow left", ``arrow right", ``bluetooth', ``close", ``fast forward", ``play", ``refresh", ``reply", ``rewind", ``send"\} -- all of which have are arrow-like and are typically pointing to the left or right.

These visually meaningful clusters translate to performance improvements similar to that seen in the CIFAR experiment, with learning efficiencies as high as 2.58 (from the hierarchies induced by task similarity). As above, the hierarchical classifiers contained $ 100 $ trees for the coarse labels and $ 100 $ trees for the fine labels with each tree having a maximum depth of $ 10 $. The flat classifiers consisted of $ 300 $ trees with a maximum depth of $ 20 $.
\label{real-data}

\section{Discussion}

Our empirical results on synthetic as well as real-world data confirm the hypothesis that assuming existence of a latent hierarchy can lead to significant improvement in down-stream classification tasks, even in scenarios where there is no known latent structure. Naturally, our performance gains have a strong dependence on the clustering algorithm as well as the embedding space over which clustering is performed. 

Further, the baseline performance in each real-world task therefore does not reflect state-of-the-art accuracy; however, our reported results indicate substantial improvement – in the case of CIFAR~100, inducing a latent structure resulted in 26.7\% absolute increase in accuracy (53.8\% classification accuracy) when trained on just 10\% of the data. Likewise for the ICONS dataset with a classification accuracy of 76.2\%, a 37.6\% absolute improvement over the baseline flat classifier (38.6\% accuracy). Note that in both datasets, the number of classes was large, hence were likely more amenable to a hierarchical representation. Similar improvements were not observed in the CheXpert dataset which consists of significantly fewer classes. Future directions of this work include empirical results on large, benchmark datasets with many classes such as ImageNet, a more detailed set of CIFAR~100 experiments utilizing a state-of-the-art baseline, as well as a theoretical basis for why the hierarchical assumptions hold in practice. 

We believe our results will encourage broad adoption of hierarchical classification in a vast majority of practical settings where the baseline representation is fixed a priori via off-the-shelf models and embeddings, and that such methods may become essential in the small sample size regime.
\label{discussion}

\clearpage
\bibliographystyle{iclr2021/iclr2021_conference}
\bibliography{biblio}

\clearpage

\label{appendix}


\end{document}